 \documentclass[final,5p,times,twocolumn]{elsarticle}

\usepackage{amssymb}

\usepackage{bm}
\usepackage{amsmath,multirow}
\usepackage{url}
\usepackage[acronym]{glossaries}
\newacronym{pcqagsv}{PCQA-GSV}{point cloud quality assessment with graph signal variations}
\newacronym{dslg}{DSLG}{diagonal scan-line graph}
\newacronym{svr}{SVR}{support vector regression}

\newacronym{ngsv}{NGSV}{normalized graph signal variations}
\newacronym{knn}{KNN}{K nearest neighbors}
\newacronym{glr}{GLR}{graph Laplacian regularization}
\newacronym{cd-sgw}{CD-SGW}{spectral graph wavelet-based color denoising}
\newacronym{3dpbs}{3DPBS}{3-dimensional patch-based similarity}
\newacronym{lbvh}{LBVH}{linear bounding volume hierarchy}
\newacronym{fgbd}{FGBD}{fast graph-based denoising}
\newacronym{slg}{SLG}{scan-line graph}
\newacronym{ne-gbp}{NE-GBP}{noise estimation using graph-based patches}
\newacronym{fslr}{FSLR}{filter selection with limited region}
\newacronym{pca}{PCA}{principal component analysis}
\newacronym{mvub}{MVUB}{Microsoft voxelized upper bodies}
\newacronym{psnr}{PSNR}{Peak signal-to-noise ratio}
\newacronym{bf-knn}{BF-KNN}{Brute-force KNN}
\newacronym{megw}{MEGW}{median estimator with graph wavelets}
\newacronym{qa}{QA}{quality assessment}
\newacronym{pcqa}{PCQA}{point cloud quality assessment}
\newacronym{nr-pcqa}{NR-PCQA}{no-reference PCQA}
\newacronym{nr}{NR}{no-reference}
\newacronym{fr-pcqa}{FR-PCQA}{full-reference PCQA}
\newacronym{rr-pcqa}{RR-PCQA}{reduced-reference PCQA}
\newacronym{icip-pcvqa}{ICIP 2023 PCVQA Grand Challenge}{ICIP 2023 point cloud visual quality assessment grand challenge}
\newacronym{frsvr}{FRSVR}{full-reference quality assessment using support vector regression}

\usepackage{color}
\usepackage{xcolor}
\usepackage[normalem]{ulem}

\journal{Signal Processing: Image Communication}

\begin{document}

\begin{frontmatter}



\title{Full reference point cloud quality assessment using support vector regression}

\author[KDDI,USC]{Ryosuke Watanabe\corref{cor1}}
\author[USC]{Shashank N. Sridhara}
\author[USC]{Haoran Hong}
\author[USC]{Eduardo Pavez}
\author[KDDI]{Keisuke Nonaka}
\author[KDDI]{Tatsuya Kobayashi}
\author[USC]{Antonio Ortega}

 \affiliation[KDDI]{organization={KDDI Research, Inc.},
             addressline={2-1-15 Ohara},
             city={Fujimino},
             postcode={356-8502},
             state={Saitama},
             country={Japan}}

 \affiliation[USC]{organization={University of Southern California},
             addressline={3650 McClintock Ave},
             city={Los Angeles},
             postcode={90089},
             state={California},
             country={United States}}

\cortext[cor1]{Corresponding author at KDDI Research,~Inc. (2-1-15 Ohara, Fujimino, Saitama, 356-8502, Japan), E-mail address: ru-watanabe@kddi.com (R. Watanabe).}

\begin{abstract}
Point clouds are a general format for representing realistic 3D objects in diverse 3D applications.
Since point clouds have large data sizes, developing efficient point cloud compression methods is crucial.
However, excessive compression leads to various distortions, which deteriorates the point cloud quality perceived by end users.
Thus, establishing reliable point cloud quality assessment (PCQA) methods is essential as a benchmark to develop efficient compression methods.
This paper presents an accurate full-reference point cloud quality assessment (FR-PCQA) method called full-reference quality assessment using support vector regression (FRSVR) for various types of degradations such as compression distortion, Gaussian noise, and down-sampling.
The proposed method demonstrates accurate PCQA by integrating five FR-based metrics covering various types of errors (e.g., considering geometric distortion, color distortion, and point count) using support vector regression (SVR).
Moreover, the proposed method achieves a superior trade-off between accuracy and calculation speed because it includes only the calculation of these five simple metrics and SVR, which can perform fast prediction.
Experimental results with three types of open datasets show that the proposed method is more accurate than conventional FR-PCQA methods.
In addition, the proposed method is faster than state-of-the-art methods that utilize complicated features such as curvature and multi-scale features.
Thus, the proposed method provides excellent performance in terms of the accuracy of PCQA and processing speed. Our method is available from \url{https://github.com/STAC-USC/FRSVR-PCQA}.
\end{abstract}




\end{frontmatter}


\section{Introduction}
\label{sec:introduction}

Point clouds are utilized to represent realistic 3D objects in a variety of 3D applications such as telepresence~\cite{telepresence}, autonomous driving~\cite{ADriving}, monitoring~\cite{monitoring}, and holographic display~\cite{Holo3DTV}.
These applications may introduce distortions during the scanning, compression, transmission, storage, and rendering processes.
These distortions may degrade the perceptual quality of point clouds.

In recent years, efficient point cloud compression and quality enhancement have been studied to mitigate the impact of distortions.
For instance, the MPEG committee is developing two standards for point cloud compression: geometry-based point cloud compression (G-PCC) and video-based point cloud compression (V-PCC)~\cite{MPEGPCC}. 
Moreover, compression methods based on deep learning have emerged as alternative approaches to improve coding efficiency~\cite{CNNPCC,GeoCNN}.
Furthermore, many methods aiming at improving the quality of point clouds have been proposed to alleviate the impact of distortions (e.g., point cloud denoising~\cite{PCDenoising2,PCDenoising}, upsampling~\cite{PCUpsampling,PCUpsampling2}, and inpainting~\cite{PCinpainting2,PCinpainting}).
Under the circumstances, \gls{pcqa} methods are important benchmarks in these studies.
Hence, establishing reliable \gls{pcqa} methods helps develop high-performance compression and quality enhancement.

In contrast to quality assessment methods for 2-D images or videos~\cite{VMAF,SSIM,MS-SSIM}, where only color information is utilized, perceptual quality of 3D content is significantly influenced by \textit{both} geometry and color distortion and their interactions. Thus, \gls{pcqa} methods should consider both types of distortions. 
Consequently, quality assessment methods developed for 2D images or videos cannot be directly extended to \gls{pcqa}.
Methods for \gls{pcqa} are categorized into \gls{fr-pcqa}, \gls{rr-pcqa}, and \gls{nr-pcqa} approaches. 
Unlike \gls{rr-pcqa} and \gls{nr-pcqa} metrics, \gls{fr-pcqa} methods provide stable assessments because they compare a distorted point cloud with the reference point cloud. 
Due to this stability, some \gls{fr-pcqa} metrics~\cite{MPEGEval,C2PError} have been employed as criteria for establishing effective compression methods in the MPEG standardization~\cite{MPEGPCC}.
While these methods~\cite{MPEGEval,C2PError} focus on point-wise errors, 
other methods that consider (i) more complex features (e.g., structural similarity~\cite{PointSSIM,MSPointSSIM,PCQM}),  
 (ii) graph similarity~\cite{GraphSIM,MSGraphSIM}, or (iii) learning-based techniques \cite{DeepFR,PointPCA+} have been proposed to improve the \gls{pcqa} accuracy.

The challenges with a conventional \gls{pcqa} method are assessment accuracy and computation speed.
Regarding \gls{pcqa} accuracy, although (i) more complex features and  (ii) graph similarity can enhance the accuracy of point-wise errors, the \gls{pcqa} accuracy is still limited. 
This is because the simple combination such as a linear combination~\cite{MSPointSSIM,PCQM} or multiplication~\cite{GraphSIM,MSGraphSIM} of multiple metrics has been introduced. 
In contrast, incorporating (iii) learning-based techniques significantly improves accuracy.
However, these advancements suffer from slow assessment speed.
In particular, computational complexity is a problem when we consider a large point cloud or point cloud video (dynamic point clouds).

This paper introduces a novel \gls{fr-pcqa} method,  \gls{frsvr}~\footnote{Source code of the proposed method is available at \url{https://github.com/STAC-USC/FRSVR-PCQA}.}.  
Since the proposed method can integrate multiple simple metrics effectively based on \gls{svr}, it achieves a superior trade-off between \gls{pcqa} accuracy and computation speed.
Our proposed \gls{frsvr} obtained the first place in the FR broad-range quality estimation results track in the \gls{icip-pcvqa}~\cite{PCVQA2023}.

\subsection{Contributions of this paper}

Our main contributions are:
\begin{enumerate}
    \item We propose an accurate \gls{fr-pcqa} by integrating five types of FR metrics using \gls{svr}. Since the five metrics cover various errors (e.g., geometry, color, point count, point-to-point, and region-to-region errors), we achieve excellent objective quality assessment accuracy. In addition, we achieve a favorable computation speed because the five metrics share several steps, such as neighborhood search and graph construction.
    \item We comprehensively evaluated accuracy and computation time on three public datasets and verified the effectiveness of our method for various data and distortion types, not only the compression distortions evaluated in the \gls{icip-pcvqa}.
    \item For further improvement of the accuracy of the proposed method, we experimented with the combination with scores other than the five types of scores mentioned above. We observed that the accuracy exceeded our \gls{icip-pcvqa} results. 
    These results are introduced in Section \ref{sec:discusssion} as extra experiments. The additional scores are obtained from PCQM~\cite{PCQM}, which is a state-of-the-art \gls{pcqa} method that utilizes a linear combination of various types of \gls{fr-pcqa} scores.
\end{enumerate}

The proposed methodology described in the first contribution was designed for the \gls{icip-pcvqa}~\cite{ICIPChallenge}, which aims to assess compression distortions.
The second and third contributions are new.

\subsection{Structure of this paper}

The remainder of this paper is organized as follows:
Section \ref{sec:related-work} provides a comprehensive review of related work in \gls{pcqa}.
Section \ref{sec:proposed-method} introduces the details of the proposed method.
Section \ref{sec:experiments} shows the experimental results that verify the effectiveness of the proposed method.
Based on the experimental results, Section \ref{sec:discusssion} shows the limitations of the proposed method and strategies for further improvement.
Finally, we conclude the paper in Section \ref{sec:conclusion}.

\begin{figure*}[t]
\centering 
   \includegraphics[width=0.90\linewidth]{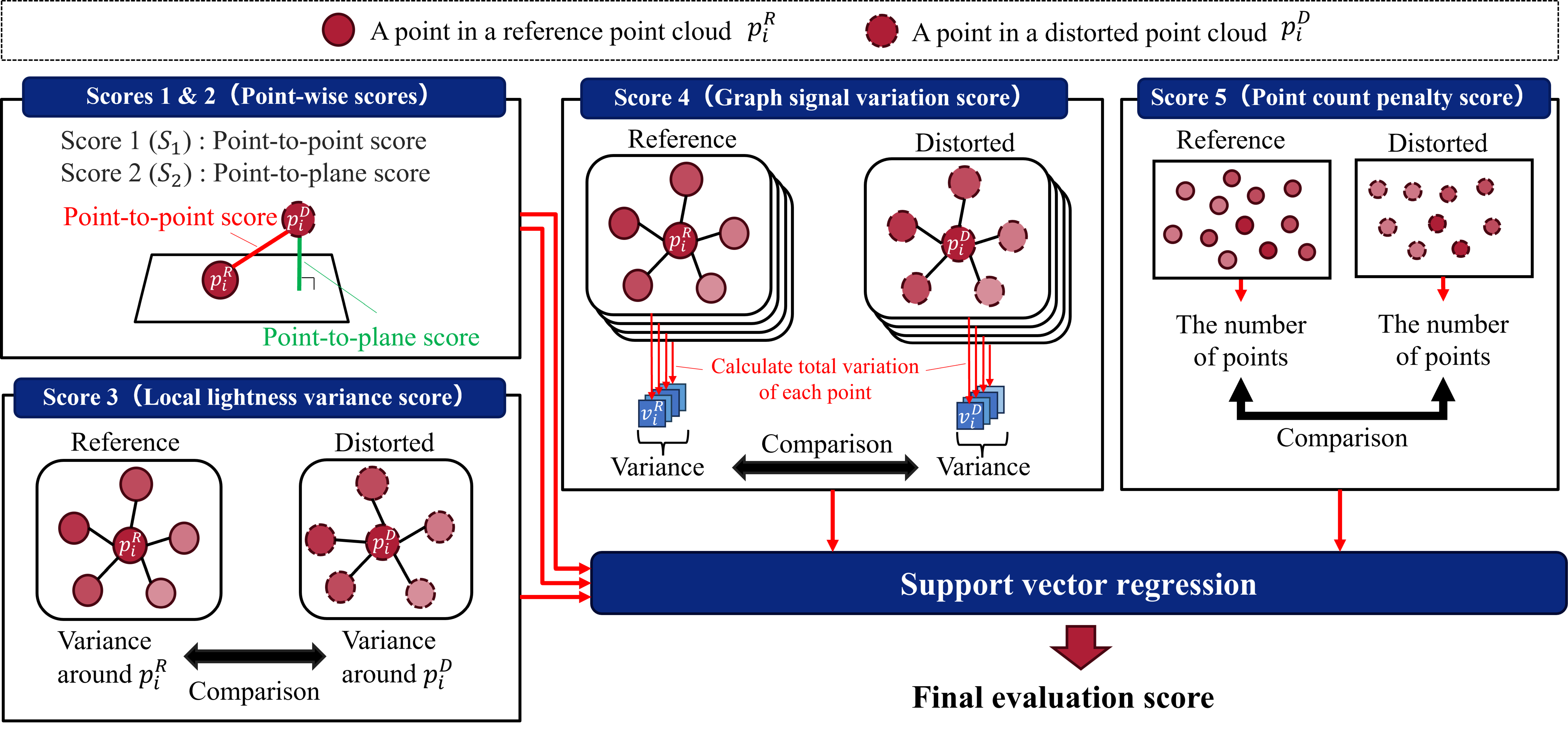}
   \caption{The calculation flow of the proposed method. The five \gls{fr-pcqa} scores are utilized to predict the final evaluation score by \gls{svr}.}
\label{fig:flow-fr}
\end{figure*}

\section{Related work}
\label{sec:related-work}

\gls{pcqa} methods are categorized into the following three classes according to whether reference point clouds are required: (1) \gls{nr-pcqa}, (2) \gls{rr-pcqa}, and (3) \gls{fr-pcqa}.

\subsection{No-reference PCQA (NR-PCQA) methods}

\gls{nr-pcqa} methods can be applied when reference point clouds are not available.
Learning-based approaches are mainstream for \gls{nr-pcqa}.
Some methods perform 3D-to-2D projection and then use NR image quality assessment based on convolutional neural networks~\cite{MVP-PCQA,CVS-PCQA,PQANet}.
As an alternative that does not require projections, methods using 3D deep neural networks~\cite{ContextualDNN,Psinet,Logistic-PCQA} and graph neural networks~\cite{GPA-Net,PCQA-GraphPoint} have been proposed.
While neural network-based approaches achieve accurate quality assessment, the processing time increases due to 3D-to-2D projection or inference with deep neural networks.
Methods using \gls{svr} have been proposed~\cite{Ext3D-NSS,3D-NSS} as alternatives to neural network techniques,  
demonstrating a favorable trade-off between computation speed and accuracy.
However, the stability and robustness of NR quality assessment can be low because the results generally depend strongly on the training data's characteristics.

\subsection{Reduced-reference PCQA (RR-PCQA) methods}

\gls{rr-pcqa} approach utilizes partial information in a reference point cloud for \gls{pcqa}.
Viola and Cesar~\cite{RR-PCQA1} propose an \gls{rr-pcqa} method using a small set of statistical features from the reference point cloud.
Liu et al.~\cite{RR-PCQA2} utilize the attribute and geometry quantization steps of point cloud compression methods to infer the point cloud quality.
Zhou et al.~\cite{RR-PCQA3} introduce an image-based \gls{rr-pcqa} method via saliency projection. The content-oriented similarity and statistical correlation calculated from saliency maps are utilized to estimate the perceptual quality of a point cloud.
Since the accuracy of \gls{rr-pcqa} depends on the extracted features, it can decrease if the features are not robust or fail to capture critical point cloud information.

\subsection{Full-reference PCQA (FR-PCQA) methods}

\gls{fr-pcqa} methods, which are the focus of this paper, necessitate a high-quality (noise-free) reference point cloud for each distorted point cloud. 
They can be classified into four groups based on (1) projection, (2) 3D geometry, (3) geometry and color, and (4) learning. Our method makes use of geometry and color metrics and combines them using a simple learning strategy.

\subsubsection{Projection-based methods}

Projection-based methods utilize conventional 2D image quality assessment techniques after projecting a 3D point cloud onto a 2D plane. 
A recent framework~\cite{Projection} predicts point cloud quality by applying 2D image quality assessment metrics (e.g., SSIM~\cite{SSIM}, MS-SSIM~\cite{MS-SSIM}, VIFP~\cite{VIFP}) to multiple projected views.
Liu et al.~\cite{IW-SSIMp} provide an attention-guided \gls{pcqa} inspired by the information content weighted-structural similarity measure~\cite{IW-SSIM}.
To enhance the perception of geometric distortion, curvature projection is utilized to extract geometric statistical features~\cite{CurvatureProjection}.
Since these approaches result in the loss of 3D information during the projection from 3D to 2D, the assessment accuracy may be occasionally degraded.
Besides, the choice of projection process and the number of viewpoints can negatively impact the accuracy of the final prediction.
\subsubsection{3D geometry-based methods}

3D geometry-based methods measure geometric distortion in point clouds.
Point-to-point error~\cite{MPEGEval} and point-to-plane error~\cite{C2PError} are popular metrics for assessing geometric distortion. 
The former measures the distance between each point of the distorted point cloud and its corresponding nearest neighbor point in the reference point cloud. The latter refers to the projection error in the normal vector direction of the nearest point in the reference point cloud.
Also, Alexiou et al.~\cite{AngularSimilarity} proposed an \gls{fr-pcqa} method using angular similarity of tangent planes among corresponding points.
These methods are constrained because they only consider geometry. 
\subsubsection{3D geometry and color-based methods}
Some \gls{fr-pcqa} methods consider both geometric and color distortion.
Geometric distortion can indirectly affect color distortion because the positional correspondence between the distorted and reference point clouds is required to compute color distortion.
Regarding color distortion, the Peak Signal-to-Noise Ratio (PSNR) is a benchmark in MPEG standardization~\cite{MPEGEval}.
As an alternative, PointSSIM~\cite{PointSSIM}, an extension of SSIM~\cite{SSIM} for point clouds, compares statistical information of small regions, such as color variance.
\gls{pcqa} accuracy is limited since these methods compare only one feature derived from color components (e.g., color itself or color variance around a point).
To improve accuracy, PCQM~\cite{PCQM} is calculated using a weighted linear combination of geometric and color distortion metrics.
Moreover, a graph-based \gls{fr-pcqa} method~\cite{GraphSIM} has been proposed to measure distortion using a graph structure constructed from point clouds.
Computation cost is high since these methods utilize complex features, such as curvature, to improve accuracy.
Furthermore, methods that perform multi-scale comparisons have also been proposed to improve the accuracy~\cite{MSPointSSIM,MSGraphSIM}.
However, these methods significantly increase calculation time.

\subsubsection{Learning-based methods}

In recent years, learning-based approaches have been introduced to obtain better accuracy.
An end-to-end deep-learning framework \cite{DeepFR} is proposed for accurate \gls{fr-pcqa}, considering both geometry and color information. 
Another approach~\cite{PointPCA+} utilizes PCA-based features to predict the evaluation score. 
While these methods demonstrate high accuracy by combining geometry and color information, 
they suffer from high computation costs~\cite{PCVQA2023}.

\section{Proposed method}
\label{sec:proposed-method}

\subsection{Overview}
\label{subsec:prop-overview}

This section introduces the proposed \gls{fr-pcqa} method, FRSVR (see Fig.~\ref{fig:flow-fr}).
In the training process, an \gls{svr} model~\cite{SVR} is trained using five types of FR scores ($S_1 \sim S_5$), namely: (1) point-to-point score $S_1$, (2) point-to-plane score $S_2$, (3) local lightness variance score $S_3$, (4) graph signal variation score $S_4$, and (5) point count penalty score $S_5$.
The $S_1$ and $S_2$ scores represent geometric distortion. 
In contrast, the $S_3$ and $S_4$ scores quantify local and global lightness differences between distorted and reference point clouds.
Since a KNN graph constructed by coordinate values is utilized to calculate the $S_3$ and $S_4$ scores, they are affected by both lightness and geometric distortions.
$S_5$ captures the difference in the number of points.
All scores are such that $S_i\in [0,1]$, with larger values indicating better quality. 
Then, \gls{svr} is used to integrate the scores related to geometry, lightness, and the number of points.

\subsection{Preliminaries}

Distorted and reference point clouds are represented as $\mathrm{P_D} = \{p^{D}_i \},$ $ i=1,...,|\mathrm{P_D}|$ and $\mathrm{P_R} = \{p^{R}_j\},$ $j=1,...,|\mathrm{P_R}|$, respectively, where $|\mathrm{P_D}|$ and $|\mathrm{P_R}|$ denote the number of points.
Point $p^{D}_i$ has 3D coordinates $\bm{g}^{D}_i \in \mathbb{R}^3$ and associated RGB color $\bm{c}^{D}_i \in \mathbb{R}^3$ (Likewise, $p^{R}_j$ has $\bm{g}^{R}_j$ and $\bm{c}^{R}_j$).
The proposed method uses the lightness (L) term of each point $l^{D}_{i} \in \mathbb{R}$ that is calculated by color conversion from RGB color space ($\bm{c}^{D}_i$) to LAB color space~\cite{LABCOLORSPACE} before calculating the scores, as is done in the conventional \gls{pcqa} method~\cite{PCQM}.

In the proposed method, we construct an undirected graph, $\mathcal{G}=(\mathcal{V},\mathcal{E})$, from the point cloud to calculate the metrics. We use the Euclidean distance; each point is connected to its $K$ nearest neighbors (KNN).
$\mathcal{V}$ and $\mathcal{E}$ indicate the sets of nodes and edges on the graph, respectively.
The points in the set $\mathcal{N}(p_j)$ are the neighbors of $p_j$.

\subsection{Point-to-point score}
\label{subsec:p2point}
The point-to-point error~\cite{MPEGEval} is 
the sum of geometric distances between every point in one point cloud and the corresponding closest point in the other.
The point-to-point error from the reference point cloud $\mathrm{P_R}$ to the distorted point cloud $\mathrm{P_D}$ is calculated as
\begin{equation}
\label{eq:p2point-from-ref-to-dist}
E^{R}_{p2point} = \frac{1}{|\mathrm{P_R}|} \sum_{\forall p^R_{i} \in \mathrm{P_R}} 
||\bm{g}^{R}_{i} - \bm{\bar{g}}^{D}_{i}||^2_2 ,
\end{equation}
where $\bm{\bar{g}}^{D}_{i} \in \mathbb{R}^3$ are the coordinates of the nearest neighbor for the point $p^{R}_{i}$ in the distorted point cloud $\mathrm{P_D}$.
Likewise, the point-to-point error from the distorted point cloud $\mathrm{P_D}$ to the reference point cloud $\mathrm{P_R}$ is calculated as
\begin{equation}
\label{eq:p2point-from-dist-to-ref}
E^{D}_{p2point} = \frac{1}{|\mathrm{P_D}|} \sum_{\forall p^D_{j} \in \mathrm{P_D}} 
||\bm{g}^{D}_{j} - \bm{\bar{g}}^{R}_{j}||^2_2.
\end{equation}
Here, $\bm{\bar{g}}^{R}_{j} \in \mathbb{R}^3$ are the coordinates of the nearest neighbor of a point $p^{D}_{j}$ in the reference point cloud $\mathrm{P_R}$.
Next, the point-to-point error $E_{p2point}$ is given as
\begin{equation}
\label{eq:p2point-score}
E_{p2point} = \frac{E^{R}_{p2point} + E^{D}_{p2point}}{2}.
\end{equation}
Finally, the score $S_1 \in [0,1]$
is given by
\begin{equation}
\label{eq:s1-score}
S_1 = \frac{1}{1+E_{p2point}}.
\end{equation}

\subsection{Point-to-plane score}
\label{subsec:p2plane}

To calculate the point-to-plane error~\cite{C2PError}, we projected the error vector $\bm{g}^{D}_{j} - \bm{\bar{g}}^{R}_{j}$ introduced in \eqref{eq:p2point-from-dist-to-ref} along the normal vector direction of a point.
Since the point-to-plane error represents the distance between a point and the surface of a point cloud, it sometimes provides a better \gls{pcqa}, i.e., closer to human perception than that obtained from point-to-point error.
In contrast, if the estimation accuracy of normal vectors is not good, the reliability of the point-to-plane error may be low compared with the point-to-point error.
Therefore, our proposed method uses both the point-to-point and point-to-plane errors.

The point-to-plane error from 
the distorted point cloud $\mathrm{P_D}$ to the reference point cloud $\mathrm{P_R}$ is given by
\begin{equation}
\label{eq:p2plane-from-dist-to-ref}
E^{D}_{p2plane} = \frac{1}{|\mathrm{P_D}|} \sum_{\forall p^{D}_j \in \mathrm{P_D}}((\bm{g}^{D}_{j} - \bm{\bar{g}}^{R}_{j}) \cdot \bm{n}^{R}_j)^2,
\end{equation}
where $\bm{n}^{R}_j$ is the normal vector corresponding to $\bm{\bar{g}}^{R}_{j}$.
$\bm{a} \cdot \bm{b}$ denotes the inner product between two vectors, $\bm{a}$ and $\bm{b}$.
The normal vectors are calculated by a conventional normal estimation method~\cite{NVEstimation}.
After that, the point-to-plane score $S_2$ is calculated by

\begin{equation}
\label{eq:s2-score}
S_2 = \frac{1}{1+E^{D}_{p2plane}}.
\end{equation}

Note that \eqref{eq:s2-score} only computes an error from the distorted to the reference point cloud, while in \eqref{eq:p2point-score}, errors in both directions are considered (distorted to reference and vice-versa). This reduces the processing time by a factor of two, while, experimentally, there was almost no difference in \gls{pcqa} accuracy between the symmetric and asymmetric metrics. 

\subsection{Local lightness variance score}
\label{subsec:local-color-variation-error}
\begin{figure}[t]
\centering \includegraphics[width=0.95\linewidth]{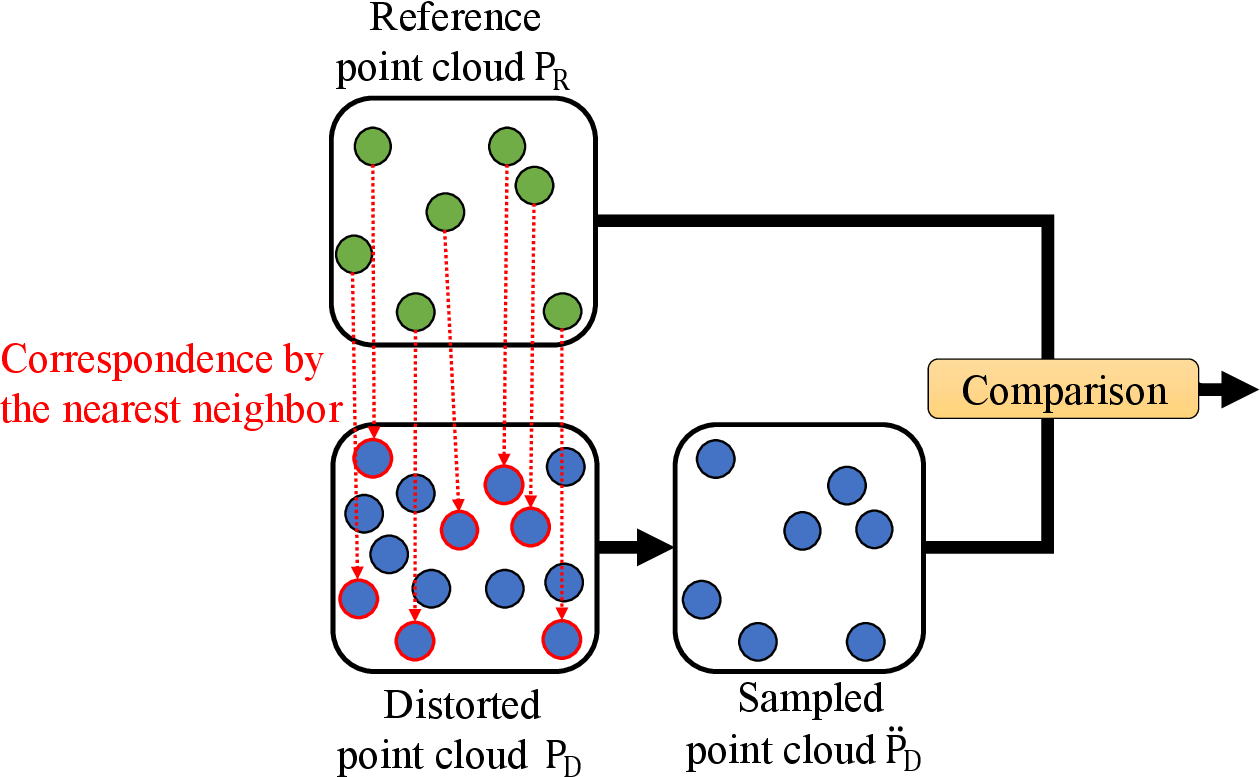}
   \caption{The sampling process to obtain a sampled point cloud $\mathrm{\ddot{P}_D}$.}
\label{fig:pc-sampling}
\end{figure}
Unlike $S_1$ and $S_2$, and similar to PointSSIM~\cite{PointSSIM}, the local lightness variance score $S_3$ addresses the distortion of lightness signals in specific regions.  
Before calculating $S_3$, a distorted point cloud $\mathrm{P_D}$ is sampled into the sampled distorted point cloud $\mathrm{\ddot{P}_D} = \{\ddot{p}^D_i\}$ as a pre-processing step to reduce computational complexity.
Figure~\ref{fig:pc-sampling} shows this sampling process.
$\ddot{p}^D_i$ is the nearest point from the query point $p^R_i$. 
The degradation on \gls{pcqa} accuracy is pretty small by adapting this sampling strategy. 
We discuss the impact of this sampling strategy in Section \ref{subsec:EX-SAMPLING}.

After the sampling, the score $S_3$ is calculated using  $\mathrm{P_R}$ and $\mathrm{\ddot{P}_D}$ as follows:

\begin{equation}
\label{eq:s3-score}
S_3 = \frac{1}{1+E_{bvar}},
\end{equation}

\begin{equation}
\label{eq:color-variance-score}
E_{bvar} = \frac{1}{|\mathrm{P_R}|}\sum_{\forall p^R_i \in \mathrm{P_R}} (\sigma^R_i - \ddot{\sigma}^{D}_i)^2,
\end{equation}
where,
\begin{equation}
\label{eq:std-r}
\sigma^R_i = \sqrt{\frac{1}{K}\sum_{\forall p^R_k \in \mathcal{N}(p^R_i)}(l^R_k - \mu^{R}_i)^2},
\end{equation}
\begin{equation}
\label{eq:std-d}
\ddot{\sigma}^D_i = \sqrt{\frac{1}{K}\sum_{\forall \ddot{p}^D_k \in \mathcal{N}(\ddot{p}^D_i)}(\ddot{l}^D_k - \ddot{\mu}^{D}_i)^2}.
\end{equation}
$\sigma^R_i$ and $\ddot{\sigma}^{D}_i$ are standard deviations of the respective lightness $l^R_k$ and $\ddot{l}^D_k$.
Besides, $\mu^R_i$ and $\ddot{\mu}^{D}_i$ are means of the K nearest neighbors from point $p^R_i$ and $\ddot{p}^D_i$, respectively.
\subsection{Graph signal variation score}
\label{subsec:graph-signal-variation-error}
The score $S_4$ is based on the difference of graph signal variations, which quantify the smoothness of signals on a graph~\cite{GTV}.
As with $S_3$, we use the sampled $\mathrm{\ddot{P}_D} = \{\ddot{p}^D_i\}$ to calculate the score.
After that, we compute the total variations of each point in $\mathrm{P_R}$ and $\mathrm{\ddot{P}_D}$, to form 
$\bm{v}^R=\{v^R_{i}\}$ and $\bm{v}^D=\{v^D_{i}\}$, with:
\begin{equation}
\label{eq:calc-s4r-score}
v^R_i =  \sum_{p^R_{k} \in \mathcal{N}(p^R_i)}  |l_i^R - l_k^R|,
\end{equation}
\begin{equation}
\label{eq:calc-s4d-score}
v^D_i =  \sum_{\ddot{p}^D_{k} \in \mathcal{N}(p^D_i)} |\ddot{l}_i^D - \ddot{l}_k^D|.
\end{equation}
$S_4$ is calculated by the difference of the standard deviations of $\bm{v}^R$ and $\bm{v}^D$ by
\begin{equation}
\label{eq:s4-score}
S_4 = \frac{1}{1+{E_{gvar}}},
\end{equation}
where
\begin{equation}
\label{eq:calc-s4-score}
E_{gvar} = \frac{|\mathrm{Std}(\bm{v}^R) - \mathrm{Std}(\bm{v}^D)|}{\mathrm{Std}(\bm{v}^R)}.
\end{equation}
$\mathrm{Std}(\bm{a})$ denotes the standard deviation of $\bm{a}$.

Note that the definition of $S_4$ differs from that in our ICIP challenge paper~\cite{ICIPChallenge}, where we used the sums of \eqref{eq:calc-s4r-score} and \eqref{eq:calc-s4d-score}. 
Noise-like errors can produce increases and decreases of the local variation of \eqref{eq:calc-s4r-score} and \eqref{eq:calc-s4d-score} for different nodes $i$. 
As an alternative, we use the standard deviations of the entries of $\bm{v}^R$ and $\bm{v}^D$ as shown in \eqref{eq:calc-s4-score} instead of \eqref{eq:calc-s4r-score} and \eqref{eq:calc-s4d-score}. Experimentally, we found that this led to better results.

\subsection{Point count penalty score}
\label{sec:proposed-method-point-count-penalty}
In general, the number of points has a significant impact on the perceived quality of a point cloud.
Specifically, when the number of points in a distorted point cloud $|\mathrm{P}_\mathrm{D}|$ is much smaller than that in the reference point cloud $|\mathrm{P}_\mathrm{R}|$, the quality of the distorted point cloud $\mathrm{P}_\mathrm{D}$ is likely to be low.
For example, if a point cloud is compressed with a small bit depth, which leads to a reduction in the number of points in the voxelization process, significant degradation of point cloud quality may be observed, as the point cloud becomes sparse.
Thus, we introduce a point cloud penalty score, $S_{5}$, that evaluates the difference in the number of points as
\begin{equation}
\label{eq:score5-point-count-penalty}
S_\mathrm{5} = \min \left(1, \frac{\mathrm{|P_D|}}{\mathrm{|P_R|}} \right).
\end{equation}
This penalizes distorted point clouds that have a small number of points as compared with the reference point cloud.
\subsection{Support vector regression (SVR)}
\label{sub-sec:svr}
In the proposed method, \gls{svr} is utilized to predict \gls{pcqa} scores.
In the training process, an SVR model is trained with the five scores $S_{1} \sim S_{5}$.
The kernel and solver are the Gaussian radial basis function kernel and Sequential Minimal Optimization (SMO)~\cite{SMO}, respectively.
\section{Experiments}
\label{sec:experiments}

\subsection{Experimental conditions}

\subsubsection{Dataset}
We utilized the following three datasets for our experiments: 1) The broad quality assessment of static point clouds in compression scenario dataset (BASICS)~\cite{BASICS}, 2) The ICIP2020 dataset (ICIP20)~\cite{ICIP20}, and 3) The waterloo point cloud database (WPC)~\cite{WPC}.
In the datasets, the mean opinion score (MOS) of the distorted point clouds obtained from a subjective quality assessment is available.
3600, 73, and 60 subjects participated in the subjective evaluation of the BASICS, ICIP20, and WPC datasets, respectively.
For all the datasets, the subjects assessed the quality of point clouds with a 2D display that shows point clouds.

\begin{itemize}
\item The \textbf{BASICS}~\cite{BASICS} dataset includes 75 references and 1498 distorted point clouds. The distorted point clouds undergo compression by V-PCC~\cite{MPEGPCC}, G-PCC~\cite{MPEGPCC}, or a deep-leaning-based compression method~\cite{GeoCNN} at different compression levels.
\item The \textbf{ICIP20}~\cite{ICIP20} dataset comprises 6 references and 90 distorted point clouds. Distortions are caused by V-PCC~\cite{MPEGPCC} or G-PCC~\cite{MPEGPCC}.
\item The \textbf{WPC}~\cite{WPC} dataset consists of 20 references and 740 distorted point clouds. This dataset includes a variety of distorted point clouds, including not only those induced by compression but also by Gaussian noise and down-sampling.
\end{itemize}

\begin{table*}[t]
\centering
\small
\caption{The test sets of the ICIP20~\cite{ICIP20} and WPC~\cite{WPC} datasets. Two and four kinds of reference data and the corresponding distorted points are utilized for testing in the ICIP20 and WPC datasets, respectively.}
\begin{tabular}{lcc}
\hline
Test set  & ICIP20~\cite{ICIP20}                    & WPC~\cite{WPC}                                                      \\ \hline
Test set 1 & {[}longdress, loot{]}      & {[}bag, banana, biscuits, cake{]}                            \\
Test set 2 & {[}redandblack, ricardo{]} & {[}cauliflower, flowerpot, glasses\_case, honeydew\_melon{]} \\
Test set 3 & {[}sarah, soldier{]}       & {[}house, litchi, mushroom, pen\_container{]}                \\
Test set 4 & N/A                        & {[}pineapple, ping-pong\_bat, puer\_tea, pumpkin{]}          \\
Test set 5 & N/A                        & {[}ship, statue, stone, tool\_box{]}                         \\ \hline
\end{tabular}
\label{tab:datasets}
\end{table*}

The proposed method was designed for the \gls{icip-pcvqa}~\cite{PCVQA2023}, which aims to assess compression distortion introduced in the BASICS dataset~\cite{BASICS}.
To confirm the effectiveness against compression distortion on another dataset, we experimented with the ICIP20 dataset in this paper.
Furthermore, we utilized the WPC dataset to evaluate the applicability to various kinds of noise.

%
\subsubsection{Evaluation Criteria}
We employed Pearson’s linear correlation coefficient (PLCC) and Spearman’s rank-order correlation coefficient (SROCC) with respect to the subjective scores to assess the \gls{pcqa} accuracy of the proposed and the conventional methods.
The PLCC and SROCC represent the prediction accuracy and the strength of prediction monotonicity of objective quality assessment metrics, respectively.
According to the recommendation for quality assessment~\cite{ITU-REC}, a four-parameter logistic regression function~\cite{logis-4} was used to calculate the correlation coefficients.
In addition, the processing time was measured using a computer equipped with an AMD Ryzen Threadripper 2970WX 24-Core processor, NVIDIA GTX 1080 Ti, and 128GB Random Access Memory. 
%
%
\subsubsection{Training and test scheme}

Since the proposed method requires training data, we performed partitioning of each dataset into training and test sets. 
The BASICS dataset was explicitly divided into training (60\%), validation (20\%), and test (20\%) sets by the authors of the dataset~\cite{BASICS}. Thus, we conducted training on the training set and evaluated the performance on the test set. 
Furthermore, we utilized the validation set for certain validations, such as parameter determination.
Both the ICIP20 and WPC datasets were segmented into three and five parts, respectively (See Table \ref{tab:datasets}). 
Each segment served as an independent test set, while the remaining data were utilized as training data. 
Following the calculation of PLCC and SROCC for each test set, the average of those was shown as the evaluation result.

\subsubsection{Implementation details}

In the proposed method, we adopted $K_3=20$ and $K_4=5$ where $K_3$ and $K_4$ indicate the number of neighbors for calculating the $S_3$ and $S_4$ scores. 
The reason is discussed in Section \ref{subsec:EX-K-choices} along with the experimental results.

As the conventional \gls{fr-pcqa} for comparison (introduced in Section \ref{subsec:EX-CONV-PROP-comparison}), we adopted point-to-point error~\cite{MPEGEval}, point-to-plane error~\cite{C2PError}, angular similatiy~\cite{AngularSimilarity}, Y-MSE~\cite{MPEGEval}, PointSSIM~\cite{PointSSIM}, PCQM~\cite{PCQM}, GraphSIM~\cite{GraphSIM}, and MSGraphSIM~\cite{MSGraphSIM}.
As for PointSSIM~\cite{PointSSIM}, though many kinds of implementations are introduced in the original paper~\cite{PointSSIM}, we introduced PointSSIM calculated by geometric signals (geom-PointSSIM) and color (color-PointSSIM).
The software of conventional \gls{fr-pcqa} methods, which is provided by the authors, does not utilize GPU acceleration.
Thus, for a fair comparison, GPU computing is not utilized for the implementation of the proposed method.

%
%

\begin{figure}[t]
\centering \includegraphics[width=0.70\linewidth]{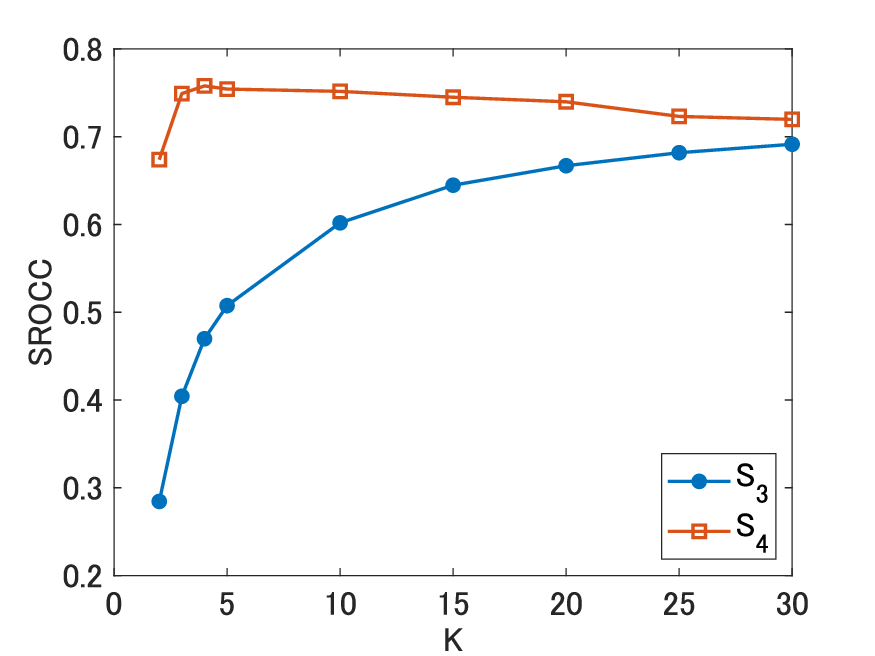}
\caption{The relationship between the SROCC and the number of neighbors ($K$) for calculating $S_3$ and $S_4$ with the BASICS dataset (validation set).}
\label{fig:graphS3S4}
\end{figure}

\begin{figure}[t]
\centering \includegraphics[width=0.65\linewidth]{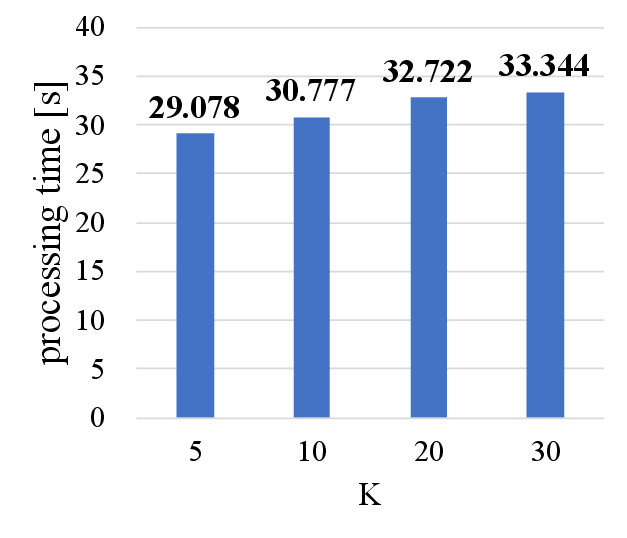}
\caption{The relationship between the average processing time per point cloud [s] and the number of neighbors ($K$) with the BASICS dataset (test set).}
\label{fig:graphTime}
\end{figure}

\subsection{Experiment 1: Effects of choices on graph construction parameter}
\label{subsec:EX-K-choices}

First, we investigated the impact of parameter choices on graph construction.
Figure \ref{fig:graphS3S4} illustrates a relationship between the SROCC and the choices of the parameter $K$ for \gls{knn} with the validation set of the BASICS dataset.
In Fig.~\ref{fig:graphS3S4}, the line labeled  ``$S_3$'' indicates the SROCC calculated from only $S_3$ score without using \gls{svr}.
As shown in Fig.~\ref{fig:graphS3S4}, since the SROCC of the $S_4$ score saturated at around $K=5$, we adopted the number of neighbors $K_{4}=5$ for the calculation of $S_4$ in the following experiments.
This is because the values of $\mathrm{Std}(\bm{v}^R)$ and $\mathrm{Std}(\bm{v}^D)$ in \eqref{eq:calc-s4-score} are strongly influenced not only by noise but also by the color changes in the original (noise-free) signals. 
The closer the two points are, the more similar their corresponding colors are in an original point cloud. 
Thus, as $K$ increases, it is the color changes of the original signal, and not the noise, that are likely to be reflected in the calculation of standard deviations, $\mathrm{Std}(\bm{v}^R)$ and $\mathrm{Std}(\bm{v}^D)$.
The error $E_{gvar}$ calculated by \eqref{eq:calc-s4-score} is influenced by the increase in $\mathrm{Std}(\bm{v}^R)$ because the denominator in \eqref{eq:calc-s4-score} is $\mathrm{Std}(\bm{v}^R)$. 
As the value of $\mathrm{Std}(\bm{v}^R)$ increases due to larger $K$, the error $E_{gvar}$ may become smaller regardless of noise.
Thus, the SROCC slightly decreases as $K$ increases.
In contrast, the accuracy of $S_3$ tends to improve as $K$ increases.
Also, Fig.~\ref{fig:graphTime} shows the relationship between the overall processing time and the choices of $K$. The processing time of the proposed method increases as $K$ increases.
Considering the balance between the accuracy and processing time, we adopted $K_3=20$ for calculating the $S_3$ score in the following experiments.
If accuracy is the primary concern, it may be effective to set larger $K$.

%
%

\begin{table}[t]
\centering
\small
\caption{The correlation coefficient of the test set of the BASICS dataset. The \textbf{bold characters} show the best performance.}
\begin{tabular}{lcc}
\hline
\multicolumn{1}{c}{\multirow{2}{*}{Score}} & \multicolumn{2}{c}{BASICS (test)} \\ \cline{2-3} 
\multicolumn{1}{c}{}                       & PLCC            & SROCC           \\ \hline
$S_1$                                     & 0.803
           & 0.738           \\
$S_2$                                     & 0.873          & 0.831           \\
$S_3$                                     & 0.686         & 0.665           \\
$S_4$                                     & 0.836          & 0.698           \\
$S_5$                                     & 0.671           & 0.235           \\ \hline
Proposed w/o $S_1$                            & 0.908           & 0.858           \\
Proposed w/o $S_2$                            & 0.909        & 0.872           \\
Proposed w/o $S_3$                            & 0.908       & 0.862           \\
Proposed w/o $S_4$                            & 0.906         & 0.877           \\
Proposed w/o $S_5$                            & 0.910           & 0.876           \\ \hline
Proposed                                   & \textbf{0.914}           & \textbf{0.878}           \\ \hline
\end{tabular}
\label{tab:EX-EACH-SCORE}
\end{table}
\begin{table}[t]
\centering
\small
\caption{The correlation coefficient of the test set of the BASICS dataset with $\hat{S_4}$ that is introduced in~\cite{ICIPChallenge}. The \textbf{bold characters} show the best performance.}
\begin{tabular}{lcc}
\hline
\multicolumn{1}{c}{\multirow{2}{*}{Score}} & \multicolumn{2}{c}{BASICS (test)} \\ \cline{2-3} 
\multicolumn{1}{c}{}                       & PLCC            & SROCC          \\ \hline
$\hat{S_4}$ \cite{ICIPChallenge}                                    & 0.818          & 0.667          \\ 
$S_4$                                 & 0.836        & 0.698           \\
\hline
Proposed with $\hat{S_4}$ \cite{ICIPChallenge}            & 0.912           &0.875           \\ 
Proposed with $S_4$                                 & \textbf{0.914}           & \textbf{0.878}          \\ \hline
\end{tabular}
\label{tab:ADD-EX-S4}
\end{table}
\begin{table}[t]
\centering
\small
\caption{Breakdown of the processing time with the BASICS dataset (test set). These figures are the average processing time across all the point clouds in the dataset.}
\begin{tabular}{lc}
\hline
Part                   & \multicolumn{1}{l}{Time [s]} \\ \hline
Color space conversion & 0.641                        \\
Graph construction     & 17.090                       \\
$S_1$ calculation         & 0.014                        \\
$S_2$ calculation         & 11.897                       \\
$S_3$ calculation         & 1.919                        \\
$S_4$ calculation         & 1.159                        \\
$S_5$ calculation         & 0.001                        \\
SVR prediction time    & 0.001                        \\ \hline
Total                  & 32.722                       \\ \hline
\end{tabular}
\label{tab:breakdown-calctime}
\end{table}

\subsection{Experiment 2: Performance of the proposed method}
Table \ref{tab:EX-EACH-SCORE} shows the PLCC and SROCC of each metric $S_1 \sim S_5$ and ablation study using the BASICS dataset's test set.
As a single metric, the $S_2$ score achieved the highest accuracy.
Since point cloud information was well integrated using \gls{svr}, the combination of five scores dramatically improved the PLCC and SROCC compared with using a single score.
Besides, ``Proposed w/o $S_i$'' in Table \ref{tab:EX-EACH-SCORE} indicates that $S_i$ score was eliminated to train an \gls{svr} model.
All the scores were necessary for the proposed method because the accuracy of the proposed method decreased when we removed one of them.
In addition, Table \ref{tab:ADD-EX-S4} compares the ${S}_4$ score calculated using \eqref{eq:calc-s4-score}  and the score $\hat{S}_4$ we utilized in our ICIP challenge paper~\cite{ICIPChallenge}, as discussed in Section 3.6.

Furthermore, Table \ref{tab:breakdown-calctime} shows a breakdown of the calculation time of the proposed method.
Before the computation of $S_1 \sim S_4$, color conversion from RGB to L color space and graph construction are carried out.
As shown in Table \ref{tab:breakdown-calctime}, the calculation time for graph construction occupied a large proportion.
In addition, since estimating the normal vectors took a long time, the processing time of $S_2$ became large.
Tables \ref{tab:EX-EACH-SCORE} and \ref{tab:breakdown-calctime} show that eliminating $S_2$ will reduce the processing time from 32.722 [s] to 20.825 (= 32.722 - 11.897) [s] with minimal degradation in the accuracy of \gls{pcqa}. Thus, 
eliminating $S_2$ would be a good option if reducing computation time is crucial.

Tables \ref{tab:EX-EACH-SCORE} and \ref{tab:breakdown-calctime} show the calculation time required by $S_5$ was less than 0.01\% of the total calculation time, while the PLCC and SROCC were improved by 0.4\% and 0.2\%, respectively.
Thus, we included $S_5$ to improve PCQA accuracy in combination with the other scores with very limited added computation.

%
%

\begin{table*}[t]
\centering
\small
\caption{The correlation coefficients and processing time [s] of the proposed and conventional \gls{fr-pcqa} methods with the three datasets. The \textbf{bold characters} show the best performance.}
\begin{tabular}{lccccccccc}
\hline
\multirow{2}{*}{Method} & \multicolumn{3}{c}{BASICS (test set)~\cite{BASICS}}   & \multicolumn{3}{c}{ICIP20~\cite{ICIP20}}   & \multicolumn{3}{c}{WPC~\cite{WPC}}                  \\ \cline{2-10} 
                        & PLCC  & SROCC & Time {[}s{]} & PLCC  & SROCC & Time {[}s{]} & PLCC        & SROCC       & Time {[}s{]} \\ \hline
    point-to-point (MSE)~\cite{MPEGEval}    & 0.793 & 0.735 & \textbf{9.701}        & 0.960 & 0.949 & \textbf{3.019}        & 0.585       & 0.566       & \textbf{6.832}        \\
point-to-plane (MSE)~\cite{C2PError}    & 0.848  & 0.799 & 49.122       &0.963 & 0.956 & 11.506       & 0.489        & 0.481       & 32.847       \\
angular similarity~\cite{AngularSimilarity}    & 0.330   & 0.306 & 39.155        & 0.650  & 0.579  & 10.462       & 0.301         & 0.319        & 27.336        \\
Y-MSE~\cite{MPEGEval}                   & 0.558 & 0.550 & 9.896        & 0.897  & 0.900 & 3.054        & 0.613       & 0.591       & 6.928        \\
PointSSIM (geom)~\cite{PointSSIM}       & 0.719  & 0.650 & 22.072       & 0.901  & 0.918 & 6.711        & 0.389       & 0.345       & 14.490       \\
PointSSIM (color)~\cite{PointSSIM}      & 0.652  & 0.620 & 23.682       &0.941  & 0.913 & 7.088        & 0.492       & 0.471       & 15.281       \\
PCQM~\cite{PCQM}                   & 0.829 & 0.739 & 256.463      & 0.952 & 0.960 & 82.801       & 0.570     & 0.550       & 214.197      \\
GraphSIM~\cite{GraphSIM}                & 0.849  & 0.773 & 613.652      & 0.943  & 0.931 & 207.532      & 0.701     & 0.691       & 382.798      \\
MSGraphSIM~\cite{MSGraphSIM}             & 0.841
 & 0.773 & 645.589      & 0.951  & 0.945 & 220.107      & 0.727  & 0.724 & 348.183  \\
Proposed (FRSVR)                & \textbf{0.914} & \textbf{0.878} & 32.722       & \textbf{0.975} & \textbf{0.965} & 10.075       & \textbf{0.819}      & \textbf{0.803}       & 20.608   \\ \hline   
\end{tabular}
\label{tab:comparison-conventional}
\end{table*}
\begin{figure}[t]
\centering \includegraphics[width=0.98\linewidth]{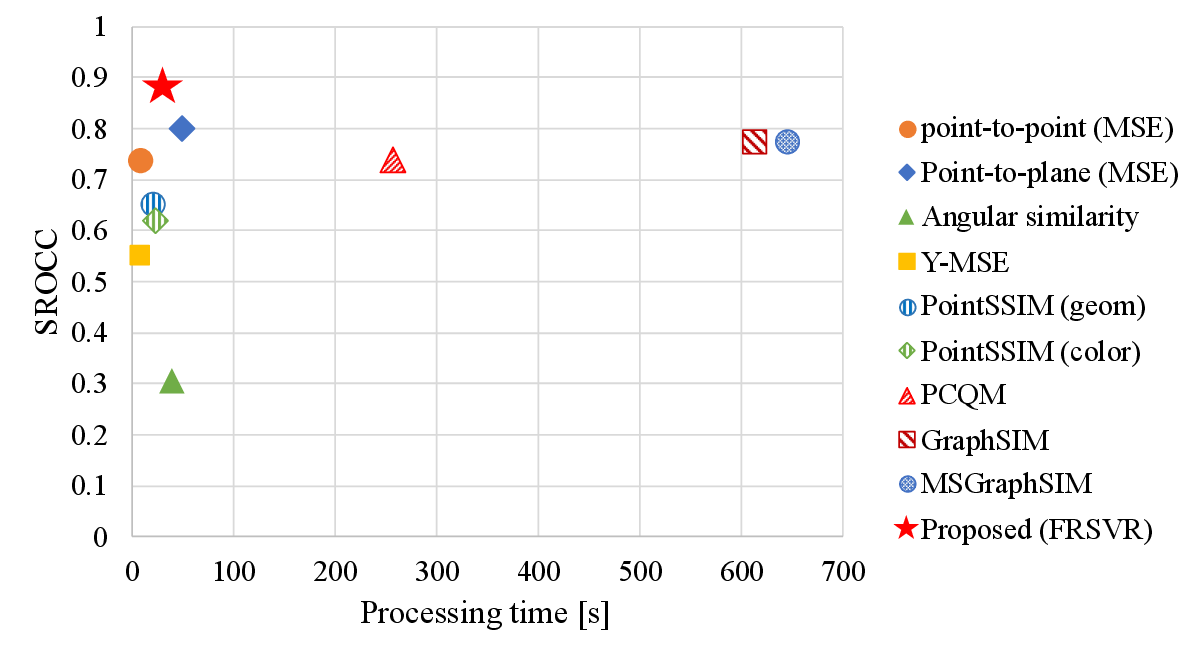}
   \caption{The SROCC and processing time [s] of the proposed method and conventional methods with the BASICS dataset (test set). The upper left of this scatter plot represents a better trade-off between the accuracy and computation speed.}
\label{fig:scatter-plot}
\end{figure}

\subsection{Experiment 3: Comparison with the conventional methods}
\label{subsec:EX-CONV-PROP-comparison}

Table \ref{tab:comparison-conventional} presents a comparative analysis of the proposed method against the conventional \gls{fr-pcqa} methods with PLCC, SROCC, and average processing time across all the point clouds.
The results demonstrate that the proposed method achieved the highest correlation coefficients across all the datasets. 
Moreover, the proposed method showed faster processing time compared with the accurate conventional methods such as point-to-plane (MSE)~\cite{C2PError}, PCQM~\cite{PCQM}, GraphSIM~\cite{GraphSIM}, and MSGraphSIM~\cite{MSGraphSIM}.
In Fig.~\ref{fig:scatter-plot}, the performance of the proposed method and conventional methods is visually depicted with a scatter plot.
These findings highlight that the proposed method achieved a superior trade-off between accuracy and computation speed.

The proposed method showed comparatively lower correlation coefficients measured with the WPC dataset in contrast with the other datasets.
Since the WPC dataset includes not only point clouds contaminated by compression errors but also those perturbed by various kinds of noise, such as downsampling and Gaussian noise, obtaining high correlation coefficients was challenging because noise has a wide variety of characteristics.
\begin{table}[t]
\centering
\small
\caption{The correlation coefficients of the BASICS dataset (test set). An \gls{svr} model is trained with either the ICIP20 or BASICS training datasets. The former indicates a cross-dataset evaluation. The \textbf{bold characters} show the best performance.}
\begin{tabular}{llcc}
\hline
Method   & Training dataset         & PLCC  & SROCC \\ \hline
Proposed & ICIP20                & 0.900  & 0.861 \\
Proposed & BASICS (training set) & \textbf{0.914} & \textbf{0.878} \\ \hline
\end{tabular}
\label{tab:gen-cap}
\end{table}
%

%
%

\subsection{Experiment 4: Generalization capability of the proposed method}
\label{subsec:EX-GENERALIZATION}

To assess the generalization capability of the proposed method, we conducted a cross-dataset evaluation. In this experiment, an \gls{svr} model was trained using all the point clouds in the ICIP20 dataset. After that, all the point clouds in the BASICS dataset (test set) were assessed by using the trained model. 
In Table \ref{tab:gen-cap}, when we utilized the ICIP20 dataset for training an \gls{svr} model, a slight degradation occurred in correlation coefficients.
However, the results were still sufficiently accurate compared with those of other conventional methods introduced in Table \ref{tab:comparison-conventional}. 
While the BASICS dataset includes point clouds compressed by a deep learning-based compression method~\cite{GeoCNN}, the ICIP20 dataset does not include them.
Consequently, differences in characteristics of compression distortion between training and test data resulted in some degree of degradation in performance.

%
%

\subsection{Experiment 5: Effect of sampling strategy in the proposed method}
\label{subsec:EX-SAMPLING}

As shown in Fig.~\ref{fig:pc-sampling}, 
we introduce point cloud sampling to mitigate computation complexity in the proposed method. Thus, we investigated the impact of the sampling on the performance. The graph construction parameters of the proposed method without sampling are the same as those of the proposed method with sampling. 
Table \ref{tab:ADD-EX5} shows that 
sampling strategy can significantly reduce the calculation time. 
For the BASICS and ICIP20 datasets, the impact on PLCC and SROCC was also very small. 
In contrast, the PLCC and SROCC of the WPC dataset decreased by sampling.
Since the WPC dataset contains a variety of noises, such as Gaussian noise, unlike the other datasets, the sampling strategy may cause an adverse effect in capturing the noise characteristic.
For example, although measuring Gaussian noise in color signals is easy in a dense point cloud, it may be difficult in a sparse point cloud because of the interference with color changes in the original signal.
If a point cloud has a variety of noise and the accuracy is a critical issue, the sampling strategy may cause a bad effect.

\begin{table*}[t]
\centering
\small
\caption{The correlation coefficients and processing time of the proposed method with or without sampling. ''Proposed (w/o sampling)'' shows the results when point clouds without sampling are utilized for the proposed method.}
\begin{tabular}{lccccccccc}
\hline
\multirow{2}{*}{Method} & \multicolumn{3}{c}{BASICS (test set)~\cite{BASICS}}   & \multicolumn{3}{c}{ICIP20~\cite{ICIP20}}   & \multicolumn{3}{c}{WPC~\cite{WPC}}                  \\ \cline{2-10} 
                        & PLCC  & SROCC & Time {[}s{]} & PLCC  & SROCC & Time {[}s{]} & PLCC        & SROCC       & Time {[}s{]} \\ \hline
Proposed (w/o sampling)             & 0.902
 & 0.873 
 & 65.039      
 & 0.974
 & \textbf{0.970}
 & 18.765  
 & \textbf{0.828}
 & \textbf{0.818}
 & 42.758  
 \\
Proposed                
& \textbf{0.914}
& \textbf{0.878}
& \textbf{32.722}    
& \textbf{0.975}
& 0.965
& \textbf{10.075}
& 0.819       
& 0.803
& \textbf{20.608}  \\ \hline   
\end{tabular}
\label{tab:ADD-EX5}
\end{table*}

\section{Discussion of the limitation and further improvement}
\label{sec:discusssion}

\subsection{Accuracy}
\label{sec:discussion-accuracy}

\begin{table*}[]
\centering
\small
\caption{The correlation coefficients of the proposed method with the additional scores derived from PCQM~\cite{PCQM}. ``Ave SROCC'' means the average of SROCCs of three datasets. The top five selected scores are listed based on the average SROCC in this table. The \textbf{bold characters} show the best performance.}
\begin{tabular}{llccccccc}
\hline
\multicolumn{1}{c}{\multirow{2}{*}{Method}} & \multicolumn{1}{c}{\multirow{2}{*}{Selected scores}} & \multicolumn{2}{c}{BASICS (test set)} & \multicolumn{2}{c}{ICIP20} & \multicolumn{2}{c}{WPC} & \multirow{2}{*}{Ave SROCC} \\ \cline{3-8}
\multicolumn{1}{c}{}                        & \multicolumn{1}{c}{}                                 & PLCC              & SROCC             & PLCC         & SROCC       & PLCC       & SROCC      &                                \\ \hline
Proposed + PCQM                             & $S_2$, $S_3$, $S_4$, $S_5$, $f_1$, $f_3$, $f_4$, $f_5$                              & 0.907             & \textbf{0.886}             & 0.975        & 0.970      & 
0.899   
& 0.898      & \textbf{0.918}                          \\
Proposed + PCQM                             & $S_2$, $S_3$, $S_4$, $S_5$, $f_1$, $f_2$, $f_3$, $f_4$, $f_5$                           & 
0.897              & 0.882             & 0.976        & 0.973       &  
\begin{tabular}{c}
\textbf{0.900}
\end{tabular}
& \textbf{0.899}      & \textbf{0.918}                          \\
Proposed + PCQM                             & $S_2$, $S_3$, $S_4$, $S_5$, $f_2$, $f_3$, $f_4$, $f_5$                               & 0.897             & 0.883             & 

0.978
       & 0.972       &   

0.899

& 0.896      & 0.917                          \\
Proposed + PCQM                             & $S_1$, $S_2$, $S_3$, $S_4$, $S_5$, $f_1$, $f_3$, $f_4$, $f_5$                            & 
0.897
                & 0.875             & \textbf{0.979}        & \textbf{0.974}       & 

\textbf{0.900}

& \textbf{0.899}      & 0.916                          \\
Proposed + PCQM                             & $S_1$, $S_2$, $S_3$, $S_4$, $S_5$, $f_1$, $f_2$, $f_3$, $f_4$, $f_5$                        & 

0.888

& 0.874             & 0.978        & \textbf{0.974}       & 
\textbf{0.900}
      & \textbf{0.899}      & 0.915                          \\ \hline
Proposed                                    & $S_1$, $S_2$, $S_3$, $S_4$, $S_5$                                       & \textbf{0.914}             & 0.878             & 
0.975

        & 0.965       & 

0.819

                & 0.803      & 0.882                          \\ \hline
\end{tabular}
\label{tab:results-with-pcqm}
\end{table*}

As shown in Table \ref{tab:comparison-conventional}, there is room for improvement, especially regarding the accuracy of the WPC dataset.
Since The proposed method integrates several \gls{fr-pcqa} scores by \gls{svr}, it can be combined with other metrics.
In this section, we introduce the combination with PCQM~\cite{PCQM}, which is a highly accurate conventional \gls{fr-pcqa} method to show further improved accuracy.
PCQM~\cite{PCQM} uses eight types of FR scores called curvature comparison $f_1$, curvature contrast $f_2$, curvature structure $f_3$, lightness comparison $f_4$, lightness contrast $f_5$, lightness structure $f_6$, chroma comparison $f_7$, and hue comparison $f_8$. 
Here, we report the results when we utilize 13 types of \gls{fr-pcqa} scores, including $f_1 \sim f_8$ and $S_1 \sim S_5$..

In this extra experiment, we verified all the subsets of 13 scores to train an \gls{svr} model.
Thus, $2^{13} - 1 = 8191$ patterns were evaluated to determine the optimal configuration.
Table \ref{tab:results-with-pcqm} shows the top five subsets (selected scores) that were evaluated based on the average SROCC of the three datasets.
When we utilized eight scores composed of $S_2$, $S_3$, $S_4$, $S_5$, $f_1$, $f_3$, $f_4$, and $f_5$, we achieved the best average SROCC (0.918). 
In particular, significant accuracy improvements were obtained for the WPC dataset with the scores derived from PCQM.
This result shows the potential for further improvement of the proposed method. 
However, PCQM has the problem of large computation time as shown in Table \ref{tab:comparison-conventional}. 
In the future, we will improve the accuracy of the proposed method by considering new features that can be calculated more efficiently.

In addition, the results shown in Table~\ref{tab:results-with-pcqm} support the includsion of $S_5$ because all of these top five subsets include $S_5$.
For example, $S_1$ was not adopted in the most accurate subset because $S_2$ shows similar assessment results to $S_1$. 
On the other hand, the $S_5$ score, which takes into account the difference in the number of points, can capture degradation that the other scores do not consider.

\subsection{Processing time}

As shown in Table \ref{tab:comparison-conventional}, while the proposed method performed faster than some of the conventional methods, the proposed method was not the fastest among all the conventional methods.
Thus, further acceleration is one of our future challenges for achieving more efficient \gls{pcqa}.
Table \ref{tab:breakdown-calctime} indicates that the computation time for graph construction is relatively large compared with other parts.
Thus, fast computation for the graph construction part is necessary.
To solve this problem, conventional fast KNN methods based on GPU implementation have been introduced~\cite{CUDA-KNN,LBVH,BN-KNN}. 
As an alternative approach, an approximation-based fast graph construction method that performs real-time processing on over 1 million points has been proposed~\cite{Authors_ICASSP2024}.
While this method~\cite{Authors_ICASSP2024} is faster than fast KNN methods~\cite{CUDA-KNN,LBVH,BN-KNN}, an inaccurate graph may be constructed and affect the accuracy of \gls{pcqa}.
In the future, we plan to assess these techniques to mitigate graph construction time and achieve faster \gls{pcqa}.
\section{Conclusion}
\label{sec:conclusion}

In this paper, we proposed an accurate and fast \gls{fr-pcqa} using \gls{svr}.
Since the proposed method integrates multiple simple metrics effectively based on  \gls{svr}, it performs a great trade-off between quality assessment accuracy and computation speed.
Our method was proposed for the \gls{icip-pcvqa}, which only focuses on quality assessments of compression distortion. 
Therefore, there are still challenges in accurate assessments for the various kinds of noise (e.g., gaussian noise, down-sampling) other than compression distortion.
While the incorporation of additional features, as discussed in Section \ref{sec:discussion-accuracy}, contributes to the improvement of accuracy, the computation time becomes significantly large.
In the future, we aim to introduce new features that are suitable for such noise and efficiently calculated to improve the proposed method.

\section*{CRediT authorship contribution statement}
\noindent \textbf{Ryosuke Watanabe}: Conceptualization, Methodology, Validation, Visualization, Formal analysis, Investigation, Software, Data curation, Writing - original draft, Writing - review \& editing.
\textbf{Shashank N. Sridhara}: Methodology, Investigation, Writing - review \& editing.
\textbf{Haoran Hong}: Methodology, Investigation, Writing - review \& editing.
\textbf{Eduardo Pavez}: Methodology, Investigation, Writing - review \& editing.
\textbf{Keisuke Nonaka}: Writing - review \& editing, Funding acquisition.
\textbf{Tatsuya Kobayashi}: Writing - review \& editing, Funding acquisition.
\textbf{Antonio Ortega}: Methodology, Investigation, Writing - review \& editing, Supervision, Project administration.

\section*{Declaration of Competing Interest}
The authors declare that they have no known competing financial interests or personal relationships that could have appeared to influence the work reported in this paper. There are no conflicts of interest regarding the publication of this study. 

\section*{Funding}
This work was supported by Ministry of Internal Affairs and Communications (MIC) of Japan (Grant no. JPJ000595).







\bibliography{JournalPaper}
\bibliographystyle{unsrt}

\end{document}